\documentclass[letterpaper, 10 pt, conference]{ieeeconf}
\IEEEoverridecommandlockouts
\usepackage{amsmath}
\usepackage{graphicx}
\usepackage{amssymb}
\usepackage{booktabs}
\usepackage{url}
\usepackage{lipsum}
\usepackage{color}
\usepackage{float}

\newif\iflongversion
\longversiontrue

\title{Deep KKL: Data-driven Output Prediction for Non-Linear Systems}

\author{Steeven Janny$^1$, Vincent Andrieu$^2$, Madiha Nadri$^2$, Christian Wolf$^3$
\thanks{$^1$Steeven Janny is with the Universit\'{e} Lyon, INSA Lyon, CNRS, LIRIS, Villeurbanne, France. E-mail: \texttt{steeven.janny@insa-lyon.fr.}}%
\thanks{$^{2}$Madiha Nadri and Vincent Andrieu are with Université Lyon, CNRS, LAGEPP, Villeurbanne, France.}%
\thanks{$^3$Christian Wolf is with Universit\'{e} Lyon, INSA Lyon, CNRS, LIRIS, Villeurbanne, France}
}
\begin{document}

\newtheorem{proposition}{Proposition}
\newtheorem{remark}{Remark}
\newtheorem{definition}{Definition}
\newtheorem{theorem}{Theorem}
\newtheorem{lemma}{Lemma}

\newcommand{\mO}{\mathcal{O}}
\newcommand{\mS}{\mathcal{S}}
\newcommand{\mY}{{Y}}
\newcommand{\mYO}{\mY_\mO}
\newcommand{\mZ}{\mathcal{Z}}
\newcommand{\RR}{\mathbb{R}}
\newcommand{\CC}{\mathbb{C}}
\newcommand{\NN}{\mathbb{N}}
\newcommand{\Cl}{\mathtt{Cl}}

\newcommand{\YY}{\mathbb Y}     

\newcommand{\fGM}{\mathfrak{f}}     
\newcommand{\hGM}{\psi}     

\maketitle
\thispagestyle{empty}
\pagestyle{empty}

\begin{abstract}
We address the problem of output prediction, ie. designing a model for autonomous nonlinear systems capable of forecasting their future observations. We first define a general framework bringing together the necessary properties for the development of such an output predictor. In particular, we look at this problem from two different viewpoints, control theory and data-driven techniques (machine learning), and try to formulate it in a consistent way, reducing the gap between the two fields. Building on this formulation and problem definition, we propose a predictor structure based on the Kazantzis-Kravaris/Luenberger (KKL) observer and we show that KKL fits well into our general framework. Finally, we propose a constructive solution for this predictor that solely relies on a small set of trajectories measured from the system. Our simulations show that our solution allows to obtain an efficient predictor over a subset of the observation space.
\end{abstract}

\section{Introduction}


\subsection{Context}
\noindent We investigate the prediction (forecasting) of future observed outputs of a non-linear dynamical system, which is not necessarily observable, and for which we have access to an initial part of the trajectory, as well as to a training set of additional representative trajectories sampled with different initial conditions. This task shares many similarities with system identification, as both problems require to design a model for a specific plant in order to represent its dynamics. Yet, in output prediction, we solely consider the system output rather than the full state representation of the system. For a long time in the literature, common solutions for this class of problem relied on explicitly modeling the physical phenomena exhibited by the dynamical system. The resulting models are then required to be as exhaustive as possible to minimize the prediction error by taking into account every part of the dynamics coming into play. 
Output predictors are central in diverse applications, like observability (e.g Kalman filtering \cite{kalman1960approach,julier2004unscented,durieu2008methodological}, Luenberger observer \cite{luenberger1964observing}) or model predictive control \cite{richalet1978model}. 

Recently, data-driven approaches based on machine learning emerged as a valuable alternative to methods based on handcrafted models for a large range of applications, where modeling is difficult, laborious or impossible. These procedures learn the dynamics directly from a set of observations of the system. In its most modern form, Deep Learning, high-capacity deep neural networks are trained from massive amounts of data, with impact on many applications in control theory by complementing classical methods \cite{ponkumar2018deep,peralez2020} or even replacing them, for instance through Deep Reinforcement Learning \cite{busoniu2018reinforcement}.
Depending on the concrete application and the amount of available data, recent work tends to demonstrate that neural networks may benefit from hybridization with more classical modeling techniques. Examples are combinations with physical models  \cite{debezenac2018deep,guen2021augmenting}, classical control techniques \cite{shi2019neural,CatchDrone2020}, or adding inductive biases to neural networks encoding domain knowledge such as projective geometry \cite{BeechingECMLPKDD2020}, path planning in graphs \cite{BeechingECCV2020}, or even objectives inspired from animal development \cite{LaetitiaMotivation2019}.

In this paper, we develop a framework for designing an output predictor for forecasting the observed output of an unknown dynamical system. 
While designed from a control theoretic point of view, it is easily transferable to methods based on Deep Learning. Moreover, under some assumptions, we show that an upper bound of the prediction error can be computed for predictors complying with our definition. 

As a use case of this general approach, we develop an output predictor based on the Kazantzis-Kravaris/Luenberger observer (KKL) \cite{kazantis1998nonlinear} for non-linear systems. Building on theoretical work \cite{andrieu2006existence, marconi2007output} and \cite{isidori2010about}, we develop a data-driven approach to compute a KKL output predictor
without any knowledge of the dynamics which generated the observations. Our method mainly relies on Deep Learning to identify relevant regularities in the training data and extracts a predictor from them. We illustrate some of the capabilities of the model across a variety of simulations. We also highlight the limitations, which are due to this constructive solution for KKL. We compare the proposition with two types of deep networks classically used in the field of machine learning for time series forecasting: \textit{Recurrent Neural Networks} (RNNs) \cite{rumelhart1986learning} and a more modern variant called \textit{Gated Recurrent Units} (GRUs) \cite{cho2014learning}.

In the same spirit, recent development around the Koopman operator \cite{lusch2018deep, rowley2009spectral} proposes to identify a transformation that projects the state of a system into an infinite dimensional latent space, in which the dynamics is fully linear, and then exploits this representation to explain the output. The Koopman operator shares with our work the idea of using Deep Learning to find a latent representation of a non-linear system from a set of observed data. Nonetheless, there are few keys differences with our contribution:
\begin{itemize}
    \item Koopman theory gives an infinite-dimensional transformation into a fully linear system. So any finite-dimensional transformation results as an approximation. By relaxing the constraint on output linearity, KKL guarantee the existence of a finite-dimensional transformation under very weak assumptions.
    \item The latent space created by the Koopman operator contains information about the full state, while our proposition requires only the \textit{observable} part of the state to be embedded. Thus, our contribution does not require neither a measurement of the complete state, nor the observability of the system.
    \item Koopman requires the mapping from the state to the latent representation and its inverse, whereas KKL only requires the identification of the inverse mapping.
    \item In contrast to our contribution, methods based on the Koopman operator do not take benefit from access to the first steps of the observed trajectory. Their predictions are solely based on the initial state of the system.
\end{itemize}
~
\vspace{-5mm}
\subsection{The output prediction problem}
\vspace{-1mm}
\noindent Consider an  unknown  dynamical system of dimension $n\in\NN$ with measured output:
\begin{equation}\label{eq:nl_sys}
    \dot x = f(x)\ , \ y=h(x)\ ,
\end{equation}
with $f:\RR^n\mapsto \RR^n$ a smooth vector field and $h:\RR^n\mapsto \RR$ a smooth observation function.
For each $x\in\RR^n$, we assume that there exists a unique solution to \eqref{eq:nl_sys}, denoted at time $t$ by $X(x_0,t)$,  with $x_0$ as initial condition.
This solution is defined for all  time (i.e. we assume forward and backward completeness). 
We introduce $\YY$, the set of all possible output functions that can be generated by this dynamical system from the set of initial conditions. Formally,
\begin{equation}
    \YY = \{y:\RR^+\mapsto \RR, \exists x_0\ ,\ y(t) = h(X(x_0,t))\}.
\end{equation}
The problem we want to solve is the following: 
\textit{Given a current time $h$ can we infer the future value of an experiment $y$ in $\YY$ given that we know $y(t)$, for $t$ in $[0,h]$ ?} Note that we may not solve this problem for all $y$ in $\YY$ but at least for those in a particular subset $\mY$ of $\YY$.


We address this problem by first defining a framework encapsulating the observation dynamics into a larger dynamical model, said \textbf{generative model} with a \textbf{contraction} property. This is similar to the idea of an \textit{internal model} \cite{francis1976internal}, as a generative model is a process  simulating the system response, with the exception that 
our definition is not necessarily motivated from control purposes. Under some assumptions, we propose an upper bound of the prediction error over time for such a model. 

In a second step, we suggest a possible solution via the Kazantzis-Kravaris/Luenberger (KKL) observer formalism. After proving the existence of a generative model under this particular form, we verify that it also respects the hypothesis required for our upper bound. To demonstrate the feasibility of this solution, and inspired by \cite{ramos2020numerical}, we design a learning algorithm to discover such KKL models. In our simulations, the KKL-based predictor exhibits remarkable forecasting capabilities,  excellent generalization and robustness to noise.

\section{Prediction via embedding into an output dependent uniform contraction }
\subsection{Uniform contraction and generating model}
\noindent Consider now a dynamical system in the form:
\begin{equation}
    \label{eq:contraction_form}
    \dot z = G(z,y) ,\ 
\end{equation}
where $z$ in $\RR^m$ and $y$ in $\RR$.
We denote by $Z(z_0,t,y)$ the solution of (\ref{eq:contraction_form}) at time $t$ initiated from an initial condition $z_0$. This solution depends only on the values of $y$ for $t$ in $[0,h]$, i.e. it is causal. 
\vspace{1mm}
\begin{definition} \label{Def1} \cite{winfried1998contraction} System (\ref{eq:contraction_form}) is said to define a \textbf{uniform exponential contraction} if there exist two positive constants $k$ and $\lambda$ s.t. for all locally integrable functions $y:\RR_+\mapsto \RR$ and all $(z_a,z_b)$ the two solutions $Z(z_a,t,y)$ and $Z(z_b,t,y)$ initiated respectively from $z_a$ and $z_b$ at $t{=}0$ satisfy:
\begin{equation}
	\label{contraction_property}
    \left|Z(z_a,t,y)-Z(z_b,t,y)\right| \leq k e^{-\lambda t} \left | z_a - z_b\right|.
\end{equation}
\end{definition}
\vspace{1mm}
\begin{remark}
We are interested in this type of dynamical systems because they \textit{forget} their initial conditions. This will be made precise in Proposition \ref{proposition_1}.
\end{remark}

Consider an autonomous system with measured output:
\begin{equation}\label{eq_GenModel}
    \dot z = g(z)\ ,\ y=\psi(z)\ ,
\end{equation}
where $g:\RR^m\rightarrow\RR^m$ and $\psi:\RR^m\rightarrow\RR$
and where the solution initiated from $z$ in $\RR^m$ and evaluated at time $t$ is denoted by $\mZ(z,t)$.
Let $\mY$ be a subset of $\YY$.
\vspace{1mm}
\begin{definition}
A \textbf{Generating Model} (GM) for $\mY$ is defined as a couple $(g,\psi)$ such that for all $y$ in $\mY$ there exists $z_{0}^y$ in $\RR^m$ such that $y(t) = \psi(\mZ(z_{0}^y,t))$.
\end{definition}
\vspace{1mm}

For instance, $(f,h)$ is a generating model for the entire set $\YY$. A generating model allows to explain an output $y$ in $\mY$ via a dynamical system. If we know the initial condition $z_0^y$ associated to $y$, future values can be predicted by integration of the GM starting from $z_0^y$.



\subsection{Prediction based on contraction and generating model}

\noindent 
We wish to predict the future of any experiments in $\mY \subset \YY$. To this end, the following definition provides two necessary conditions.
\begin{definition}
\label{def_output_predictor}
An \textbf{Output Predictor} for $\mY\subset \YY$ is defined as a couple $(G, \psi)$ such as 
\begin{itemize}
    \item $\dot z = G(z,y)$ is a uniform exponential contraction with parameter $(k,\lambda)$ as in Definition \ref{Def1};
    \item the couple $(g,\psi)$ with $g(z) = G(z,\psi(z))$ is a generating model for $\mY$. 
\end{itemize}
\end{definition}
\vspace{1mm}
\noindent The behavior of an output predictor is outlined in Figure \ref{fig:kklmodel}. Let $h$ be the number of known timesteps of $y$ and $p$ the number of predicted timesteps. For an output $y\in \mY$, we note $z_0^y$ the exact initial condition such that $\psi(Z(z_0^y,t,y)) = y(t)$ and $z_0$ the (random) initial condition used in the predictor. The prediction is decomposed into three steps:
\begin{figure}[t]
    \vspace{2mm}
    \centering
    \includegraphics[width=\columnwidth]{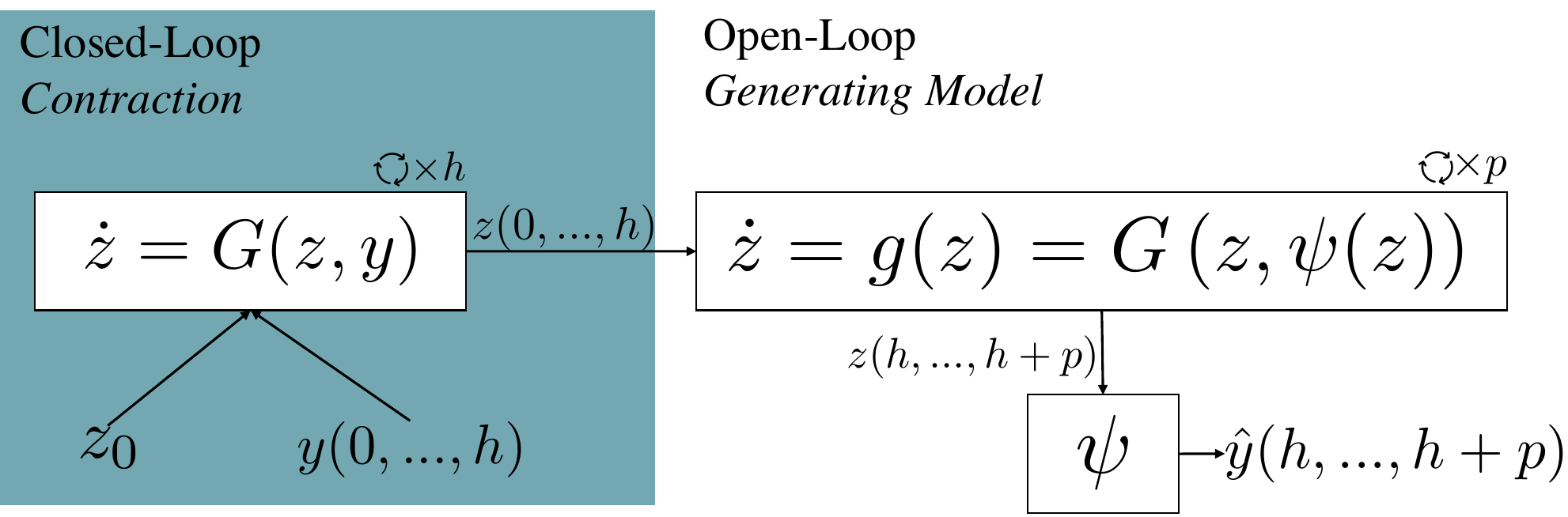}
    \caption{Computation graph for output predictors. The known part of the observation $y(t)$ for $t<h$ is used to make the latent state $Z(z_0, t, y)$ converge to $Z(z_0^y, t, y)$. 
    During the prediction step, we open the loop and let the autonomous system $\dot z = g(z)$ perform the prediction.}
    \label{fig:kklmodel}
    \vspace{-5mm}
\end{figure}
\begin{enumerate}
    \item First, the known part of the observation $y(t), t\in [0, h]$ is combined with the contraction property so that $Z(z_0, t, y)$ gets close to $Z(z^y_0, t, y)$. This is the \textbf{closed-loop} behavior of the predictor.
    \item Then, the autonomous dynamical model $\dot z = g(z)$ produces predictions in the latent space $z(t), \, t\in [h, h+p]$. We refer to this behavior as \textbf{open-loop}, since the real observation $y$ is not used as a feedback.
    \item Finally, the predicted latent state variables $z(t)$ are input to $\psi$ to compute the output $\hat y(t) = \psi(z(t))$.
\end{enumerate}




\noindent
Furthermore, if the dynamics of the latent representation $g$, and the map $\psi$ are Lipschitz, one can compute an upper bound of the prediction error due to an error on the initial condition $z_0$.
\vspace{1mm}
\begin{proposition}
\label{proposition_1}
Assume there exist $G:\RR^m\times\RR\rightarrow\RR^m$ and $\psi:\RR^m\rightarrow\RR$, both $C^1$, such that $(G, \psi)$ defines an output predictor for $\mY$ with:
    \begin{equation}
            \left|\frac{\partial g}{\partial z}(z)\right| \leq L_1\ ,\ \left|\frac{\partial \psi}{\partial z}(z) \right| \leq L_2\ ,
    \end{equation}
with $g(z) = G(z, \psi(z))$, $L_1$ and $L_2$ in $\RR^+$, then for all experiments $y\in \mY$, known in the time interval $[0,h]$, the prediction $\hat y$ at the prediction horizon $p>0$ is given as:
\begin{equation}\label{eq_yp}
    \hat y(h+p) = \psi(\mZ(Z(0,h,y),p) )\ ,
\end{equation}
and satisfies
\begin{equation}
    |\hat y(h+p) - y(h+p)| \leq k L_2 e^{-\lambda h + L_1 p}|z_0^y| \ .
\end{equation}
\end{proposition}
\vspace{1mm}
\iflongversion
\noindent
The proof of proposition \ref{proposition_1} is detailed in Appendix \ref{proof_prop1}.
\fi
\vspace{1mm}
\begin{remark}
The prediction mismatch is upper-bounded by a term, which has the following properties:
\begin{itemize}
    \item As the prediction horizon $p$ increases, the prediction error grows as well. This growth is exponential and depends mainly on the Lipschitz constant of $g$ denoted $L_1$.
    \item As $h$ increases, we obtain more information on the output before predicting. For each fixed prediction horizon, the upper-bound exponentially goes to zero for increasing $p$.
\end{itemize}
\end{remark}

\section{A possible solution via KKL}

\subsection{KKL as an output predictor}
\noindent In what follows, we derive the KKL observer structure to build an output predictor in the sense of Definition \ref{def_output_predictor}. For the sake of following mathematical consideration, the state space is reduced to a compact subset $\mO\subset \RR^n$, and we assume that it is invariant along the dynamics, ie.  for all $x_0$ in $\mO$:
$$
X(x_0,t)\in \mO\ , \forall t\, \in \RR.
$$
%
We introduce $\mYO\subset \YY$, the set of output functions that can be generated by this dynamical system when restricting $x_0$ to be in $\mO$:
\begin{equation}
    \mYO = \left\{y:\RR^+\mapsto \RR, \exists x_0 \in\mO\ ,\ y(t) = h\left(X(x_0,t)\right)\right\}\ .
\end{equation}
Inspired by the KKL observers, see \cite{kazantis1998nonlinear} or \cite{andrieu2006existence}, we consider the particular case in which the contracting model given in \eqref{eq:contraction_form} is defined on $\RR^m$ for some $m\in \NN$ and is in the form:
\begin{equation}
    \label{eq:kkl}
    G(z,y) = Az +by\ ,
\end{equation}
with $A\in \RR^{m\times m}$ a Hurwitz matrix and $b\in \RR^m$ such that $(A,b)$ is a controllable pair. The dynamical model \eqref{eq:contraction_form} with $G$ defined in \eqref{eq:kkl}, trivially defines a uniform contraction since for all $(z_a, z_b) \in \RR^m\times\RR^m$ and a given $y\in \YY$:
\begin{equation}
    \label{eq:contraction_kkl}
    |Z(z_a, t, y)-Z(z_b, t, y)| = e^{At}|z_a-z_b|\ .
\end{equation}
Since $A$ is Hurwitz, it yields the existence of $k$ and $\lambda$ such that \eqref{contraction_property} holds. 
To show that this formalism also defines a GM, we need to find $A$, $b$ and a function $\psi$ such that $\dot z = Az +b\psi(z)$ generates the output. With the use of Proposition 1, 2 and 3 from \cite{marconi2007output}, we have the following statement:
\vspace{1mm}
\begin{theorem}
\label{main_th}
With $m=2n+2$, there exist a Hurwitz matrix $A$ and a vector $b$ with $(A,b)$ controllable and a continuous mapping $\psi:\RR^m \mapsto \RR$ such that with $G$ defined in \eqref{eq:kkl}, $(G,\psi)$ defines an output predictor for $\mYO$.
\end{theorem}
\vspace{1mm}
 
 Thus, this result confirms that a linear contraction in the form \eqref{eq:kkl} may define an output predictor. 
 \iflongversion
 The proof of Theorem \ref{main_th} is given in the Appendix.
 \fi
 
 \vspace{1mm}
\begin{remark}
Going through the proof, it turns out that almost any complex couple $(A, b)$ of dimension $m'=n+1$ can be chosen to prove the existence of $\psi$, as long as $A$ is Hurwitz and $(A,b)$ controllable. One can readily extend the $m'$-dimensional complex case to our $m$-dimensional real equation by choosing $m=2m'$.
\end{remark}

\subsection{Lipschitz KKL predictor}
\noindent
The bounds on the prediction error obtained  in Proposition \ref{proposition_1}
depend on the Lipschitz constants of $\psi$ and $g$ where $g(z) = Az + b\psi(z)$.
However, the mapping $\psi$ obtained from Theorem \ref{main_th} may not be globally Lipschitz. 
In \cite{isidori2010about}  some sufficient conditions have been obtained to construct a globally Lipschitz mapping $\psi$ based on some geometric observability assumptions.
Inspired by the result obtained in \cite{andrieu2014convergence} it can be shown that when the dynamical system to predict is observable, a global Lipschitz mapping $\psi$ may be obtained. Consequently, Proposition \ref{proposition_1} may be employed.
\vspace{1mm}
\begin{proposition}
\label{prop:psi_lipshitzObs}
Assume that $h$ is a globally Lipschitz mapping.
Assume moreover that the following two observability conditions are satisfied.
\begin{itemize}
    \item \textit{Backward Distinguishability:} for all $(x_1,x_2)$ in $\mO^2$ such that $x_1\neq x_2$, there exists $t\leq 0$ such that $h(X(x_1,t)) \neq h(X(x_2,t))$.
    \item \textit{Backward Infinitesimal Distinguishability:}
    for all $(x,v)$ in $\mO\times\RR^n$ such that $v\neq 0$, there exists $t\leq 0$ such that 
    $$
    \frac{\partial h(X(x,t))}{\partial x}v \neq 0
    $$
    \end{itemize}
then 
there exist a Hurwitz matrix $A$, a vector $b$ with $(A,b)$ controllable, a mapping $\psi:\RR^m \mapsto \RR$ and a positive real number $L_2$ such that 
\begin{enumerate}
    \item with $G$ defined in \eqref{eq:kkl} $(G,\psi)$ defines an output predictor for $\mYO$;
\item the function $\psi$ has bounded derivative. i.e. 
    $$
    \left|\frac{\partial \psi}{\partial z}(z)\right|\leq L_2\ ,\ \forall z\in \RR^m;
    $$
\item the conclusion of Proposition \ref{proposition_1} holds with $L_1 = \|A\| + \|b\|L_2$.
\end{enumerate}
\end{proposition}
\noindent
Actually, the assumptions of the former proposition can be weakened by assuming that there exists an (unknown) change of coordinates, such that in such a coordinate
system \eqref{eq:nl_sys} takes a triangular form
\begin{equation}
    \label{eq:decomposition}
    \left\{\begin{array}{l}
         \dot x_1 = f_1(x_1)  \\
         \dot x_2 = f_2(x_1, x_2)
    \end{array}\right., \quad y = h(x_1) \ ,
\end{equation}
for which the couple $(f_1,h)$ satisfies the observability assumptions of the proposition.
In that case, the former proposition may be applied. Assuming the existence of this change of coordinates is very similar to the assumptions made in \cite{isidori2010about} to obtain that this mapping is globally Lipschitz.

\section{Learning $\psi$ with deep networks}
\noindent
In what follows, we propose a constructive method to find $\psi$ based on Deep Learning. We suppose to have access to two different types of data: (i) during a training phase, we have access to a representative training set of sample trajectories $\mY_D \subset \YY$ to learn $\psi_\theta(z)$, where we now have made explicit in the notation the dependency of $\psi$ on learned parameters $\theta$; (ii) for each experiment, as described in the previous sections, we have access to the initial output trajectory $y(t)\in \mY$ for $t<h$, and are required to forecast the future output up to time $h+p$ (where $p$ is the prediction horizon).


\subsection{Modeling $\psi_\theta$}
\noindent
We model function $\psi_\theta$ as a \textit{Multilayer Perceptron} (MLP) where $\theta$ in $\Theta\subset \RR^q$ is the set of parameters to be learned. This class of models is known to have universal approximation power under mild conditions either for infinitely wide \cite{UApproxWide1991} or infinitely deep (ie. layered) \cite{UApproxDepthNeurips2017} model architectures, and they also have the advantage that methods exist to limit the Lipschitz constants of the class of learned functions  (see \cite{bartlett2017spectrally, scaman2018lipschitz} for example). 


Since the previous section proves the existence of $\psi\theta$ regardless of the choice of $(A,b)$ (as long as $A$ is Hurwitz), we decided to learn $A$ freely and fix $b = (1\, ... 1)$, which reduces the number of degrees of freedom of the model. All parameters are trained by gradient descent to minimize:
\begin{multline}
    (\theta, A) =\arg \min_{\theta, A} \sum_{y\in \mY_D} \sum_{t=0}^{h+p} \|y(t) - \psi_\theta(z(t))\|^2 \\
    \text{subject to } \dot z(t) = \left\{\begin{array}{l}
         Az(t) + by(t) \text{ if } t<h \\
         Az(t) + b\psi_\theta(z(t)) \text{ else} 
    \end{array}\right.
\end{multline}
For the sake of implementation simplicity, we used a discrete formulation of the dynamics for our simulations. 
\subsection{Data-sets and baselines}
\noindent
We compare our proposition to two classical types of deep neural networks designed for time series, namely \textit{Recurrent Neural Networks} (RNNs) \cite{rumelhart1986learning}
    \begin{equation}
        z_{t+1} = \tanh(W_1z_t + W_2y_t+b)
\end{equation}
and \textit{Gated Recurrent Units} (GRUs) \cite{cho2014learning}
\begin{align}
        \openup -1\jot
        \begin{split}
                 r_{t+1} &= \sigma(W_{r1}y_t + W_{r2}z_{t} + b_r)\\
                 x_{t+1} &= \sigma(W_{x1}y_t + W_{x2}z_t + b_x) \\
                 n_{t+1} &= \tanh(W_{n1}y_t + r_{t+1}*(W_{n2}z_t + b_{n2}) + b_{n1})\\
                 z_{t+1} &= (1-x_{t+1})* n_{t+1} + x_{t+1}* z_t.
        \end{split}
\end{align}
\noindent where $\sigma(x)= (1+e^{-x})^{-1}$ is the sigmoid function and $*$ is the Hadamard product. These models contain inductive biases in the form of a recurrent memory vector $z_t$, which allows to propagate hidden state over time $t$. In other words, they define latent dynamical systems $z_{t+1} = G(z_t, y_t)$. The function $\psi_\theta$ has the same structure for each of the two variants. 

To our knowledge, no proof exists that RNNs and GRUs define proper output predictors in the sense of Definition \ref{def_output_predictor}. Depending on the learned matrix $W_1$, the function learned by the RNN may define a contraction (since $\tanh$ is monotonic), but there is no rigorous proof that $\psi$ exists for this formalism.

We evaluate our proposition on four different problems that exhibits chaotic behavior: \textit{Van Der Pol} oscillator \cite{vanderpol1926relaxation}, \textit{Lorenz} attractor \cite{lorenz1963deterministic}, \textit{Lotka-Volterra} predation equations \cite{volterra1931lecons} and a \textit{Mean Field} model \cite{noack2003hierarchy} for a fluid flow past a cylinder. \iflongversion Appendix \ref{training_details} provides details about these systems. \fi 

\section{Numerical Simulations}
\begin{figure*}[t!]
\vspace{2mm}
    \centering
    \includegraphics[width=0.9\textwidth]{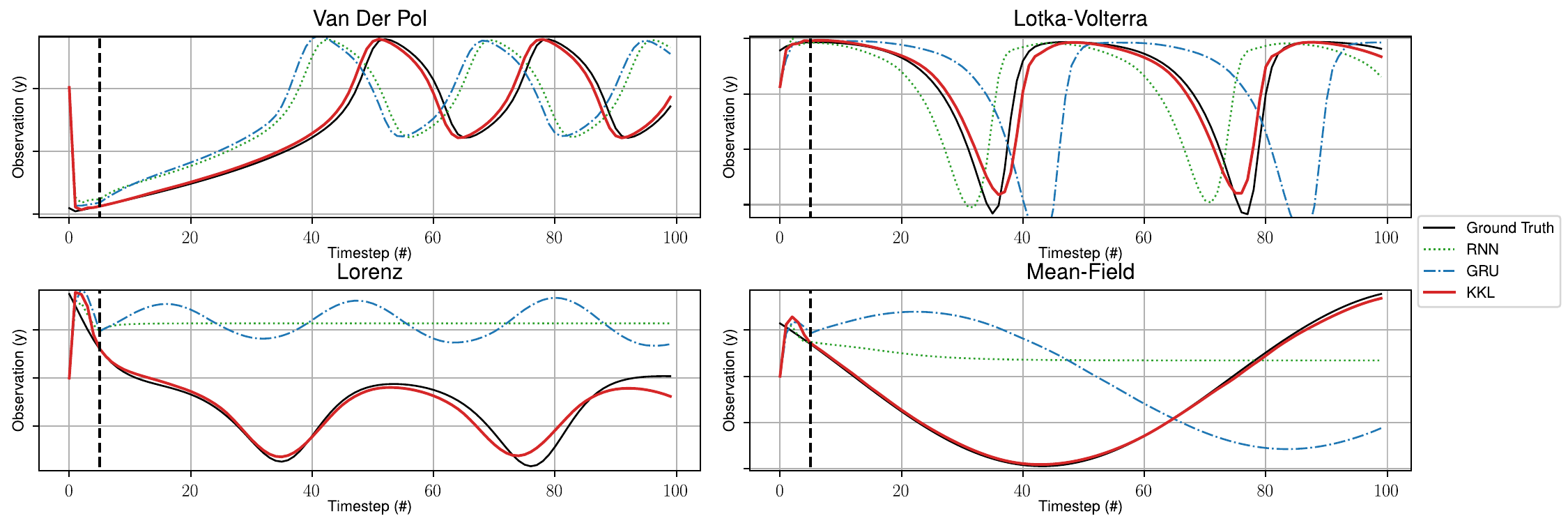}
    \vspace{-5mm}
    \caption{Demonstration of output prediction on four non-linear systems. The $t=5$ first time step (before the vertical black line) were used in the closed loop behavior of each models, then the open-loop predicts the $p=95$ following measurements}
    \label{fig:demo}
\end{figure*}

\subsection{Global performances}

\begin{table}[t]
\vspace{2mm}
    \centering
    \begin{tabular}{l|c|c|c}
                            &   RNN  &   GRU  & KKL    \\ \toprule
            Van Der Pol     & 0.0057 & 0.0343 & 0.\textbf{0013} \\
            Lotka-Volterra  & \textbf{0.0885} & 0.1780 & 0.1064 \\
            Lorenz          & 0.0441 & 0.0480 & \textbf{0.0262} \\
            Mean-Field      & 0.2254 & 0.2044 & \textbf{0.0012} \\\bottomrule
    \end{tabular}
    \caption{MSE on testing set with $h=5$ and $p=95$. The accuracy of Deep KKL is at least equal to those of the classic GRU and RNN.}
    \label{tab:mse}
    \vspace{-1cm}
\end{table}

\noindent Table \ref{tab:mse} reports the Mean Squared Error (MSE) on prediction for each model on all four datasets, namely:
\begin{equation}
    \mathcal{L}_\text{MSE} = \frac{1}{Np}\sum_{y\in \mY_T} \sum_{t=h}^{h+p}(y(t) - \hat y(t))^2
\end{equation}

\noindent where $\mY_T$ is the test set of trajectories, of cardinality $N$. To evaluate the temporal generalization capacities of all models, they were evaluated on a more difficult task than the one they were trained on. They were trained on predicting $p{=}25$ future measurements by exploiting $h{=}25$ previous measurements. However, during testing, the MSE of Table \ref{tab:mse} was calculated over $p{=}95$ predictions  after having seen only $h{=}5$ initial time steps. The results show that KKL generalizes efficiently over this broader horizon, despite the drastic decrease in the amount of data supplied as input (see Figure \ref{fig:demo}).

On our test systems, the accuracy of Deep KKL is at least equal to those of the  classic GRU and RNN, in spite of its inherent simplicity. By our simulations, we show that Deep KKL is efficient for output prediction on systems of small dimension, while offering a structure more suitable for the elaboration of guarantees. Nevertheless, in practice, the RNN and GRU deep models are rarely used in their simple form, and are generally stacked, i.e. multi-layered, where one layer takes as input the state of the previous layer. We do not claim, that on systems with very complex dynamics (stochasticity / uncertainty, large dimensions, strong non-linearity, etc.) Deep KKL will be competitive with more complex and expressive models (eg. \cite{debezenac2018deep, Baradel_2020_ICLR}). However, in our examples, Deep KKL takes advantage of its simpler structure and manages to perform better. This seems to indicate that for systems of moderate complexity, the use of high-capacity deep models does not seem to be a guarantee of better results. 

\iflongversion
\subsection{Noise Robustness}
\begin{figure}
    \centering
    \includegraphics[width=0.9\columnwidth]{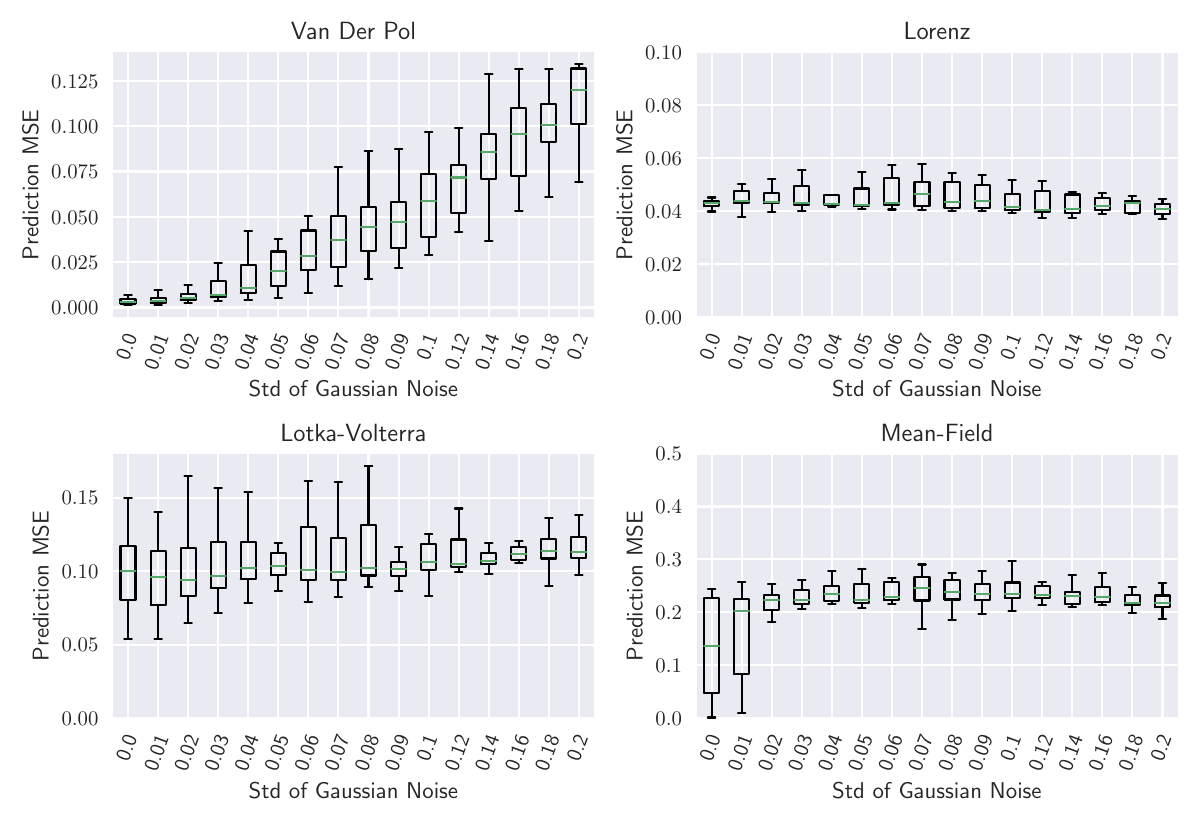}
    \vspace{-4mm}
    \caption{Boxplot of MSE on the test set $Y_T$ according to the amount of noise added during training. Observation measurements lie in $[-1, 1]$. Deep KKL is capable to deal with a reasonable amount of noise in the training data.}
    \label{fig:noise}
    \vspace{-5mm}
\end{figure}
\noindent
In an experimental setup, measurements are inevitably disturbed by noise and errors, either due to mechanical disturbances on the systems or electronic noise associated to the measurement, etc. We decided to evaluate these settings by training our model on noisy observations. In practice, we altered the measured output $y\in Y_D$ with  Gaussian noise of zero mean and varying standard deviation.

Figure \ref{fig:noise} shows the evolution of prediction error made by Deep KKL as a function of the amount of noise added to the training set. Our proposed method is still able to learn with a reasonable amount of noise on the training data. 
\fi
\subsection{Limitations due to Learning}
\noindent
On top of the initialization error detailed in Proposition \ref{proposition_1}, using Deep Learning implies another source of error due to the fact that for a given $\theta$ in $\Theta$ the estimation $\psi_\theta$ is merely an approximation of the true $\psi$ on $\mY$, which leads to errors in the open-loop phase of the prediction process.
The universal approximation theorem of neural networks \cite{csaji2001approximation} guarantees that if we allow the set of necessary parameters to be arbitrary large, then for an arbitrary choice of a constant $\delta>0$, there exists a set of parameters $\theta$ in $\Theta$ such that 
  \begin{equation}
    \label{eq:error_deeplearning}
 |\psi(z) - \psi_\theta(z)| \leq \delta\ , \quad z\in \RR^m .
\end{equation}
The evaluation of the constant bound $\delta>0$ is difficult, since we do not have access to the ground truth $\psi$. The errors $ |\psi(z) - \psi_\theta(z)|$ can have multiple reasons, and we will here ignore aspects of learnability \cite{Valiant1984PAC}, 
and concentrate on how a given error obtained by $\psi_\theta$ impacts the prediction error over time.
We formalize this as the following proposition.



\begin{figure}
    \vspace{2mm}
    \centering
    \includegraphics[width=0.49\columnwidth]{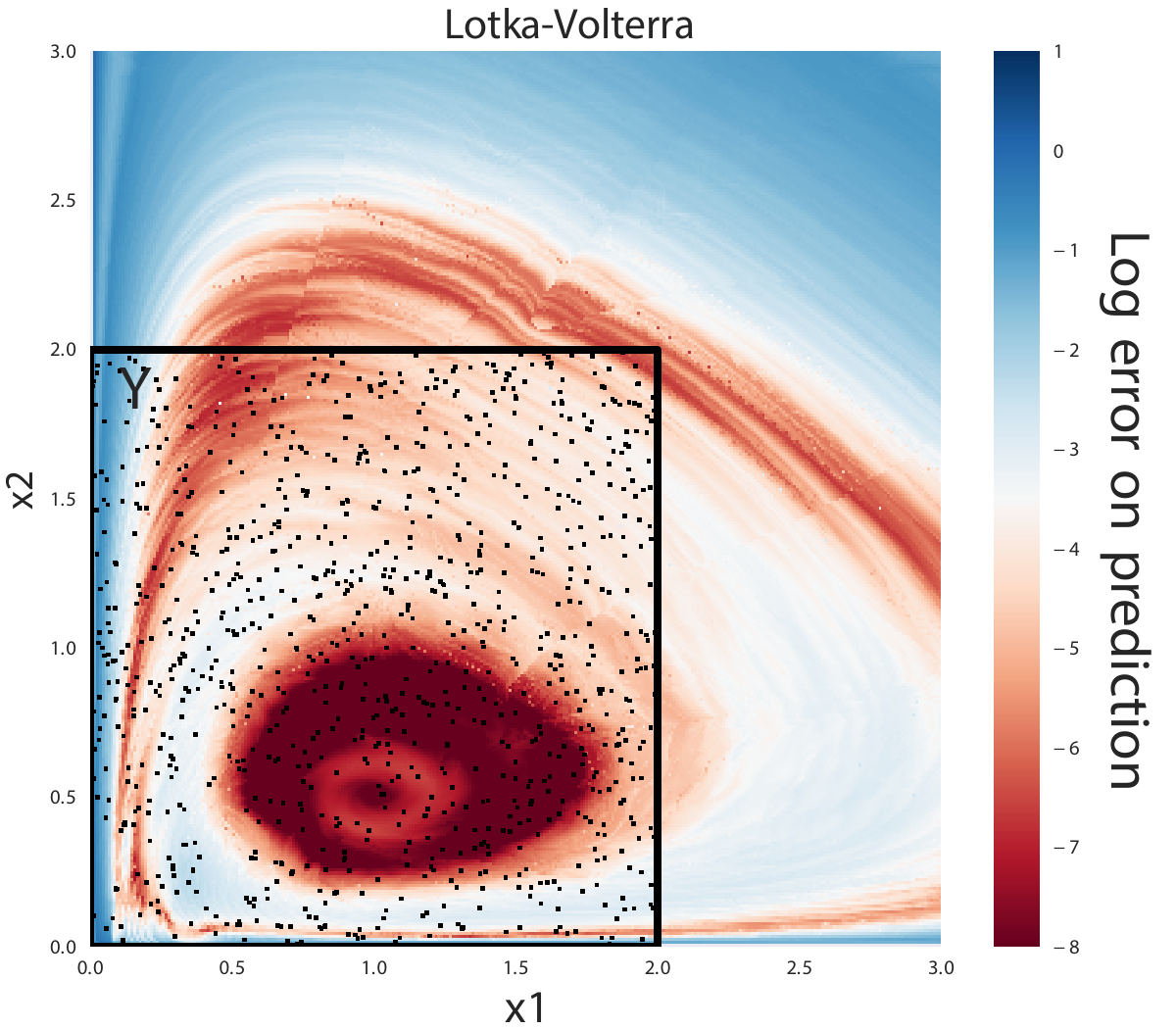}
    \includegraphics[width=0.49\columnwidth]{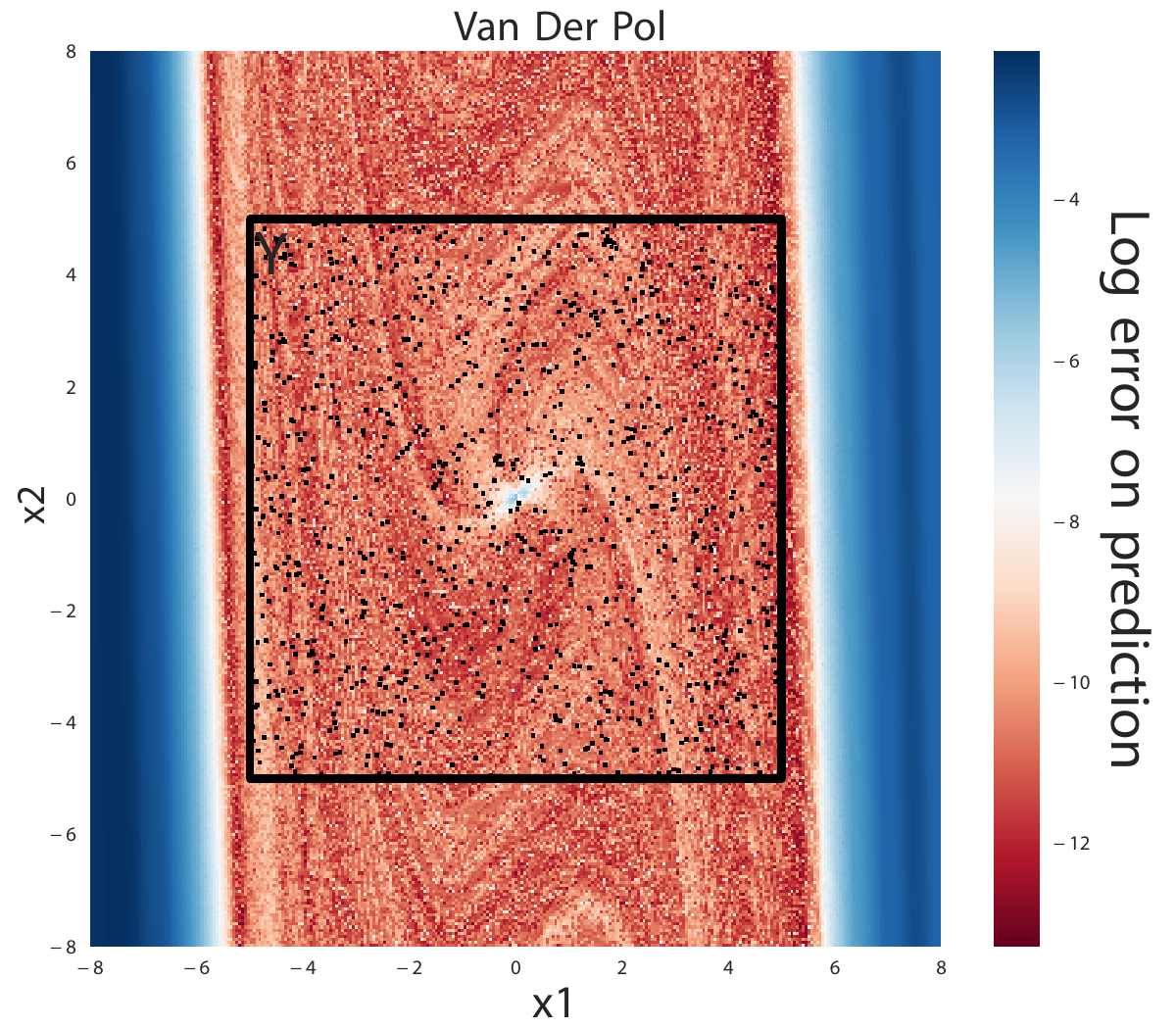}
    \caption{Generalization on unseen domain $\YY$ of Deep KKL for \textit{Van Der Pol} and \textit{Lotka-Volterra} equations. Each dot represents log-MSE on a trajectory starting from the corresponding initial condition $(x_1\, x_2)$. The black square represents the domain of the training set, training trajectories are black dots. }
    \label{fig:generalisation}
    \vspace{-5mm}
\end{figure}

\vspace{1mm}
\begin{proposition}
\label{prop:upperbound_dl}
Consider $\mY\subset\YY$.
Assume that $(A,b,\psi)$ exists such that $(G,\psi)$ with $G$ defines in \eqref{eq:kkl} is a KKL output predictor for $\mY$. Assume moreover that:
    \begin{equation}\label{Liph}
             \left|\frac{\partial \psi}{\partial z}(z) \right| \leq L_2\ .
    \end{equation}
    and that $\theta$ in $\Theta$ and $\delta>0$ satisfy \eqref{eq:error_deeplearning}.
Then for all experiments $y\in \mY$, known in the time interval $[0,h]$, a prediction $\hat y_{\theta}$ at the prediction horizon $p>0$  given as:
\begin{equation}
    \hat y_{\theta}(h+p) = \psi_\theta(\mZ_{\theta}(Z(0,h,y),p) )\ ,
\end{equation}
where $\mZ_{\theta}(z_0,p)$ is the solution initiated from $z_0$ at time $p$ of
\begin{equation}\label{eq_Sysztheta}
\dot z_\theta = Az_\theta + b \psi_\theta(z_\theta)\ ,
\end{equation}
satisfies
\begin{multline}
    |y(h+p) - \hat y_{\theta}(h+p)| \leq k L_2 e^{-\lambda h + L_1 p}|z_0^y| \\+
        \delta \left(\sqrt{e^{L_3p}-1} + 1\right) \ ,
\end{multline}
for some positive numbers $k, \lambda, L_1, L_3$ depending on $L_2$, $A$ and $b$.
\end{proposition}
\iflongversion
The proof for this proposition is given in appendix \ref{sec:proof_upperbound_dl}.
\fi




We complete this theoretical analysis by an experimental evaluation, in particular visualization of the generalization capabilities of our model. A central question in machine learning is how a model can generalize from the data it has seen during training, and thus how it performs on unseen data. Of particular interest is the distinction between ID (in-distribution) and OOD (out-of-distribution) cases, the latter describing the performance of the model on samples taken from large parameter spaces unseen during training. 
We explore this question and visualize the behavior of Deep KKL on a larger domain than the set from which the training trajectories have been sampled.

In Figure \ref{fig:generalisation}, we compute the Log-MSE of Deep KKL on a grid of trajectories from the \textit{Van der Pol} oscillator and \textit{Lotka-Volterra} equations. For each point on the heat-map, we generate the true trajectories from the corresponding initial condition $x(t=0) = (x_1\quad x_2)^T$ by integrating the corresponding ODE. Then, we use Deep KKL to predict the output of this system and compare the trajectories. The black square represents the set from which the trajectories in $\mY_D$ were sampled.

There is evidence for excellent in-distribution generalization, as Deep KKL generalizes well inside the set parameter space covered by $Y_D$, of course beyond the samples of $Y_D$ themselves. 
However, we observe limited, but not full OOD generalization, with failure cases when certain parameters are extended beyond the range seen during training.
\section{Conclusion}
\noindent
We have proposed a theoretical framework for predicting the output of dynamical systems, making it possible to easily define a device capable of representing the dynamics of the observations, and resting solely on two properties. Our proposal is illustrated in a KKL observer combined with learning a solution on a subspace of the observation space with deep neural networks.
Our simulations validate our theoretical results, and demonstrate that Deep KKL is  capable of representing the dynamics of chaotic systems of low dimension. However, the use of a learning methods inevitably generates a certain error in the estimates of $\psi$. Therefore, we proposed a quantification of the effect of this error on the predictions over time.

Future work will address learnability and sample complexity and explore the derivation of sufficient conditions on the training set $Y_D$ and on the working set $Y$ required for low estimation error $\delta$.

\iflongversion
\section{APPENDIX}
\subsection{Proof of Proposition \ref{proposition_1}}
\label{proof_prop1}

\noindent Note that 
\begin{equation}
    Z(z_0^y,h,y) = \mZ(z_0^y,h) \ .
\end{equation}
Hence, with the contraction property (\ref{Def1}), it gives~:
\begin{equation}
|Z(0,h,y) - \mZ(z_0^y,h)|\leq k e^{-\lambda h} |z_0^y|\ .
\end{equation}
Due to the Lipschitz property, it yields for all $(z_a,z_b)$ and all $p\geq 0$
\begin{equation}
|\mZ(z_a,p) - \mZ(z_b,p)|\leq e^{L_1 p} |z_a-z_b|\ .
\end{equation}
Setting $z_a =Z(0,h,y) $ and $z_b=\mZ(z_0^y,h)$, the former inequality becomes
\begin{multline*}
|\mZ(Z(0,h,y),p) - \mZ(z_0^y,h+p)|\\
\begin{aligned}
&\leq e^{L_1p} |Z(0,h,y) - \mZ(z_0^y,h)|\ ,\\
& \leq k e^{-\lambda h+L_1p}|z_0^y|\ .
\end{aligned}
 \end{multline*}
Since $(g,\psi)$ is a generating model, and since (\ref{Liph}) holds, it yields
\begin{multline}
|\hat y(t+p) - y(t+p)|\\ 
\begin{aligned}
&=|\psi(\mZ(Z(0,h,y),p)) - \psi(\mZ(z_0^y,h+p))|,\\
&\leq  L_2 k e^{-\lambda h+L_1p}|z_0^y|\ .
\end{aligned}
\end{multline}
\begin{flushright}
    \QED
\end{flushright}

\subsection{Proof of Theorem \ref{main_th}}
\label{proof_main_th}
\noindent
Theorem \ref{main_th} mostly relies on the results obtained in \cite{andrieu2006existence} in the context of observer designs and \cite{marconi2007output} in the context of output regulation. 
The proof of this statement relies on the existence of a $C^1$ function $T:\mO \mapsto \RR^m$ mapping $x$ to $z$ which satisfies the differential equation :
\begin{equation}
    \label{ap:eqdiff}
    L_f T(x) = AT(x) + bh(x) \quad \forall x\in \mO\ ,
\end{equation}
\noindent where $L_f T$ is the Lie derivative of $T$ along $f$. The functions $\psi$ and $T$ need to satisfy the equality
\begin{equation}\label{eq:psih}
\psi(T(x)) = h(x)\quad \forall x\in \mO\ .
\end{equation}


\noindent
Given a Hurwitz matrix $A$, as shown in \cite{andrieu2006existence}, the following function $T$
\begin{equation}\label{ap:eqdiff2}
    T(x) = \int_{-\infty}^0 e^{-At}bh(X(x, t)) dt\ ,
\end{equation}
is well defined for $x$ in $\mO$ and satisfies \eqref{ap:eqdiff}.
It can be shown that $T$ is $C^1$ if the eigenvalues of $A$ are smaller than a specific value depending on the Lipschitz constant of $f$. 
The proof of this results is detailed in \cite{andrieu2006existence} (see Theorem 2.4). 
To find a function $\psi$ such that \eqref{eq:psih} is satisfied, we need to ensure that $T$ contains enough information to represents the observation $y$. This requirement can be expressed as a pseudo-injectivity with regards to $h$ :
\begin{equation}
    \label{ap:injectivity}
    \forall (x_1, x_2)\in \mO \quad T(x_1) = T(x_2) \Rightarrow h(x_1) = h(x_2) .
\end{equation}
It is shown in \cite[Proposition 2]{marconi2007output} that this condition is satisfied provided $m = 2(n+1)$ and $A$ is the real representation of a Hurwitz diagonal matrix.
%
Finally, 
\cite[Proposition 3]{marconi2007output} states the existence of $\psi$.


In conclusion, if the dimension of $z\in \RR^m$ is greater or equal to $m=2n+2$, then there exists a continuous function $\psi:\RR^m \mapsto \RR$ such that for any experiments $y$ in $\mYO$, there exists $z_0^y$ such that:

\begin{equation}
    \begin{array}{l}
        \dot z = Az + by \quad z(0) = z_0^y \\
        \psi\big(Z(z_0^y, t, y)\big) = y(t) \quad \forall t
    \end{array}
\end{equation}

\subsection{Proof of Proposition \ref{prop:psi_lipshitzObs}}
\noindent
The proof of Proposition \ref{prop:psi_lipshitzObs} relies mostly on the results presented in \cite{andrieu2014convergence}.
We follow the steps of the proof of Theorem \ref{main_th}.
However, it is shown in \cite[Proposition 3.5]{andrieu2014convergence} and \cite[Proposition 3.6]{andrieu2014convergence}
that if $m=2n+2$, there exist $(A,b)$ such that the function $T$ given in \eqref{ap:eqdiff2} is injective and full rank in $\mO$. Employing \cite[Lemma 3.2]{andrieu2014convergence}, we obtain the existence of a positive real number $L_T$ such that 
\begin{equation}
    \label{ap:inverse_lips}
    L_T |T(x_1)-T(x_2)| \geq |x_1-x_2|, \quad \forall (x_1, x_2)\in \mO\ .
\end{equation}
Hence, denoting $L_h$ the Lipschitz constant of $h$, for all $(z_1,z_2)$ in $T(\mO)^2$, it yields
\begin{align*}
    |h(T^{-1}(z_1)) - h(T^{-1}(z_2))| &\leq L_h|T^{-1}(z_1) - T^{-1}(z_2)|\ ,\\
    &\leq L_h L_T|z_1-z_2|\ . 
\end{align*}
Defining $\psi$ as a global Lipschitz extension of $h \circ T^{-1}$ to $\RR^m$ yields the first and second part with $L_2=L_hL_T$ of the proposition.
The third part of the Proposition is simply obtained by noticing that with $g(z) = Az + b\psi(z)$,
$$
\left|\frac{\partial g}{\partial z}(z)\right| = \left|A + b\frac{\partial \psi}{\partial z}(z)\right| \leq |A| + |b|L_2.
$$
\subsection{Training and Architecture details}
\label{training_details}
\subsubsection{Creating the dataset} We used the following systems to evaluate our proposition. $\delta t$ is the final sampling time, and $\mathcal{D}$ the set from where the initial conditions were sampled.

\begin{itemize}
    \item \textbf{Van der Pol Oscillator} \cite{vanderpol1926relaxation} : 
    $\delta t = 0.25$ and $x_0\in \mathcal{D} = [-5, 5]^2$
    \begin{equation}
        \left\{\begin{array}{ll}
          \dot x_1 &= x_2 \\
        \dot x_2 &= (1-x_1^2)x_2-x_1
        \end{array}\right.
    \end{equation}

    \item \textbf{Lorenz Attractor} \cite{lorenz1963deterministic} 
    $\delta t = 0.02$ and $\mathcal{D} = [-20, 20]\times[-1, 1]^2$
    \begin{equation}
        \left\{\begin{array}{ll}
        \dot x_1 &= 10(x_2 - x_1) \\
        \dot x_2 &= 24x_1 - x_2 - x_1x_3 \\
        \dot x_3 &= x_1x_2 - \frac{8}{3} x_3
        \end{array}\right.
    \end{equation}
    
    \item \textbf{Lotka-Volterra Equations} \cite{volterra1931lecons} 
    $\delta t = 0.25$ and $\mathcal{D} = [0, 2]^2$
    \begin{equation}
        \left\{\begin{array}{ll}
        \dot x_1 &= x_1(\frac{2}{3} - \frac{3}{4}x_2) \\
        \dot x_2 &= x_2 (x_1 - 1)
     \end{array}\right.
    \end{equation}
    
    \item \textbf{Mean-Field} \cite{noack2003hierarchy} 
    We set $\delta t = 0.05$ and sample the initial conditions such that $x_1 = r\cos \theta$, $x_2=r\sin \theta$ and $x_3 = x_1^2+x_2^2$ with $r\in[0, 1.1]$ and $\theta \in [0, 2\pi]$, as suggested by \cite{lusch2018deep}.
    \begin{equation}
        \left\{\begin{array}{ll}
        \dot x_1 &= 0.1x_1 - x_2 - 0.1x_1x_3 \\
        \dot x_2 &= x_1 + 0.1x_2 - 0.1x_2x_3 \\
        \dot x_3 &= -10(x_3 - x_1^2 - x_2^2)
     \end{array}\right.
    \end{equation}
\end{itemize}
\par For each model, we tried to predict the observation $y=h(x) = x_1$. We used 1000 trajectories for the training set and 200 for the validation and testing set respectively. These trajectories are generated by solving the differential equation numerically using RK4 solver with a resolution $10\times$ superior than the final sampling. Finally, the observations have been re-scaled so that the training set lies between $-1$ and $1$.

\subsubsection{Training details} $\psi_\theta$ is an MLP with 3 hidden layers of 128 neurons each. We used ReLU activation functions. Canonically, the dimension of the latent space is equal to $2n+1$ where $n$ is the dimension of the system. Each model is trained with Adam optimizer for 800 epochs, with 64 trajectories per batches.

The learning rate is set to $10^{-4}$. During training, the model takes as input the $h=25$ first time steps of the output and outputs the $p=25$ following time step. Hyper-parameters were optimized over the validation set. For testing, we reduced $h$ to $5$ time steps, and increased $p$ to 95.

\subsection{Proof of Proposition \ref{prop:upperbound_dl}}
\label{sec:proof_upperbound_dl}
\noindent                                  
The idea of the proof is to compare $\hat y_{\theta}$ obtained from $\psi_\theta$ with  the prediction $\hat y$ defined in \eqref{eq_yp} obtained employing the nominal mapping $\psi$.
Note that 
\begin{equation}
    |\psi(z) - \psi_\theta(z_\theta)| 
    \leq  |\psi(z) -\psi(z_\theta)| + |\psi(z_\theta) - \psi_\theta(z_\theta)|\ .
\end{equation}
With \eqref{eq:error_deeplearning} and knowing that $\psi$ is $L_2$-Lipschitz
\begin{equation}
    |\psi(z) - \psi_\theta(z_\theta)| \leq L_2|z-z_\theta| + \delta\ .
\end{equation}
On the other hand, $A$ being Hurwitz, there exist $P$ a positive definite matrix and $\lambda>0$ such that
$$
AP + A^TP \leq -2\lambda P\ .
$$
For two vectors $(u,v)$ in $\RR^m$, let us denote $\langle u, v\rangle_P = u^\top P v$ and $\|u\|_P=u^TPu$.
Along the solutions of the system \eqref{eq_Sysztheta} and \eqref{eq_GenModel} with $g(z) = Az + b\psi(z)$ it yields
\begin{multline}
   \frac{\partial}{\partial t}  \|z-z_\theta\|_P^2 = (z-z_\theta)^T(AP+A^TP)(z-z_\theta) \\
    + 2 \langle z-z_\theta, b(\psi(z)-\psi_\theta(z_\theta))\rangle_P\ .
\end{multline}
Since $\langle u,v\rangle_P \leq \frac{ 2\lambda\|u\|_P}{2} + \frac{ \|v\|_P}{4\lambda}$, it gives
\begin{multline}
    \frac{\partial}{\partial t}  \|z-z_\theta\|_P^2 \leq -2\lambda  \|z-z_\theta\|_P^2 + \Big( \lambda\|z - z_\theta\|_P^2 \\
    + \frac{\|b\|_P^2}{\lambda}|\psi(z) - \psi_\theta(z_\theta)|^2\Big)\ .
\end{multline}
Again with \eqref{eq:error_deeplearning} and $\psi$ Lipschitz, it yields 
\begin{equation}
    \frac{\partial}{\partial t} \|z-z_\theta\|_P^2 \leq \frac{\|b\|_P^2}{2\lambda} \Big( L_2\|z-z_\theta\|_P^2 + \delta\Big)\ .
\end{equation}
With Gr\"onwall inequality, it yields,
\begin{equation}
    \|\mZ(z,p) - \mZ_\theta(z,p)\|_P^2 \leq \frac{\delta^2}{L^2_2} \left(e^{\frac{L^2_2\|b\|_P^2}{2\lambda}p} - 1\right)\ ,\forall (z,p).
\end{equation}
This implies with $\hat y$ defined in \eqref{eq_yp}~:
$$
|\hat y(h+p) - \hat y_{\theta}(t+p)|\leq \delta \left(\sqrt{e^{\frac{L^2_2\|b\|_P^2}{2\lambda}h}-1} + 1\right).
$$
However, 
\begin{multline}
    |y(h+p) - \hat y_{\theta}(h+p)| \leq \\
    |y(h+p) - \hat y(h+p)| + |\hat y(h+p) - \hat y_{\theta}(h+p)| ,
    \end{multline}
    and employing Proposition \ref{proposition_1}
it finally implies
\begin{multline}
    |y(h+p) - \hat y_{\theta}(h+p)| \leq k L_2 e^{-\lambda h + L_1 p}|z_0^y| \\+
        \delta \left(\sqrt{e^{\frac{L^2_2\|b\|_P^2}{2\lambda}p}-1} + 1\right)\ ,
\end{multline}
where $k$ is obtained from $P$ and $L_1 =  \|A\|+L_2\|b\|$ and $L_3=\frac{L^2_2\|b\|_P^2}{2\lambda}$.
This concludes the proof.
\begin{flushright}
    \QED
\end{flushright}

\fi

\bibliographystyle{IEEEtran}
\bibliography{biblio.bib}

\end{document}